\documentclass{article}

\usepackage[preprint]{acl}

\usepackage{amsmath,amsfonts,bm}

\def\eqref#1{equation~\ref{#1}}

\def\1{\bm{1}}

\DeclareMathAlphabet{\mathsfit}{\encodingdefault}{\sfdefault}{m}{sl}
\SetMathAlphabet{\mathsfit}{bold}{\encodingdefault}{\sfdefault}{bx}{n}

\usepackage{times}
\usepackage{latexsym}
\usepackage[T1]{fontenc}
\usepackage[utf8]{inputenc}
\usepackage{inconsolata}
\usepackage{hyperref}
\usepackage{url}
\usepackage{booktabs}
\usepackage{tabularx}
\usepackage{enumitem}
\usepackage{xcolor}
\usepackage{xspace}
\usepackage{microtype}
\usepackage{graphicx}

\newcommand{\system}{\textsc{RAC}\xspace}
\newcommand{\code}[1]{\texttt{#1}}
\newcommand{\leaderlogo}[1]{\raisebox{-0.22ex}{\includegraphics[height=1.75ex]{#1}}}
\definecolor{diffgreen}{HTML}{1A7F37}
\newcommand{\diffcell}[1]{\textcolor{diffgreen}{#1}}
\definecolor{diffred}{HTML}{CF222E}
\newcommand{\diffcellneg}[1]{\textcolor{diffred}{#1}}

\title{\system: A Relational Agent Contract for Constraint-Governed Engineering Tasks}
\author{
\textbf{Kaiyuan Zheng}\textsuperscript{$\clubsuit$} \;\;\;
\textbf{Fengqing Jiang}\textsuperscript{$\clubsuit$} \;\;\;
\textbf{Yichen Feng}\textsuperscript{$\clubsuit$} \;\;\;
\textbf{Zubin Aaron Assadian}\textsuperscript{$\clubsuit$} \;\;\;
\textbf{Yuetai Li}\textsuperscript{$\clubsuit$} \\
\textbf{Bhaskar Ramasubramanian}\textsuperscript{$\diamondsuit$} \;\;\;
\textbf{Basel Alomair}\textsuperscript{$\clubsuit,\spadesuit$} \;\;\;
\textbf{Luyao Niu}\textsuperscript{$\clubsuit$} \;\;\;
\textbf{Radha Poovendran}\textsuperscript{$\clubsuit$} \\
  \textsuperscript{$\clubsuit$}University of Washington \;
  \textsuperscript{$\diamondsuit$}Western Washington University \\
  \textsuperscript{$\spadesuit$}King Abdulaziz City for Science and Technology (KACST) \\
  {\texttt{\{kaiyuan5,fqjiang,yfeng42,zassad,yuetaili,alomair,luyaoniu,rp3\}@uw.edu}, \texttt{ramasub@wwu.edu}}
}

\begin{document}

\maketitle
\begin{abstract}
Engineering agents must produce artifacts that satisfy exact, coupled constraints.
Existing solutions often require agents to make semantic decisions and also derive consequences such as precise geometry or timing from those decisions, leading to failures in constraint-governed engineering tasks. 
In this paper, we introduce \system, a harness-agnostic contract for the action and feedback layer of engineering agents. 
It specifies authoritative observations, semantic actions, tool-based realization, independent verification, and repair evidence. The agent decides which relations an artifact should satisfy; specialized tools derive their concrete consequences. On PCB placement, routing, and repair tasks, \system achieves the highest clean-pass rate for all nine evaluated models, averaging 90.9\%, compared with 74.0\% for the strongest baseline, requirements-only delegation. In job-shop scheduling, direct relational actions achieve 96.3\% quality pass versus 59.6\% for code-mediated delivery with access to the same scheduling backend. Our evaluations on 141 embedded-code tasks across four microcontroller identify the action contract as a source of agent capability: preserving semantic control while delegating mechanically derivable outputs improves the reliability of constrained artifacts.

\end{abstract}

\section{Introduction}
\label{sec:introduction}

AI agents can write code, invoke tools, and operate engineering software~\citep{jimenez2024swe,sweagent2024,ChenPCBAgent2025,xu2025embedagent}. Producing a valid engineering artifact, however, requires exact constraints to hold together. 
A printed circuit board (PCB) must satisfy placement legality, clearance, connectivity, and task-specific relations such as a decoupling capacitor sitting next to the chip it serves; a manufacturing schedule must respect operation precedence and exclusive machine use.
We call these \emph{constraint-governed engineering tasks}.
Validity of artifacts targeting these tasks depends on coupled, checkable constraints.

At present, however, agents still struggle accomplishing constraint-governed engineering tasks. 
Our results suggest that \emph{judgment–derivation coupling}, where agents both make semantic decisions and reproduce their metric or hardware-specific consequences, contributes substantially to these failures.
Our insight to address such failures is that a semantic decision in these tasks typically is a \emph{relation} among named objects, such as ``C1 close to U1'' or ``operation $o_2$ ends before $o_7$ starts,'' and that relations have three properties the derived quantities lack.
First, a relation is not derivable from the engineering context: choosing it requires interpreting the requirement.
Second, once a relation is fixed, its concrete consequences follow mechanically from the context and can be computed by a solver.
Third, a relation is checkable on the realized artifact independently of whoever produced it.
Relations are therefore the natural currency of the agent's action: they capture exactly the judgment the agent must supply and nothing the tools can derive.

\begin{figure*}[t]
\centering
\includegraphics[width=\textwidth]{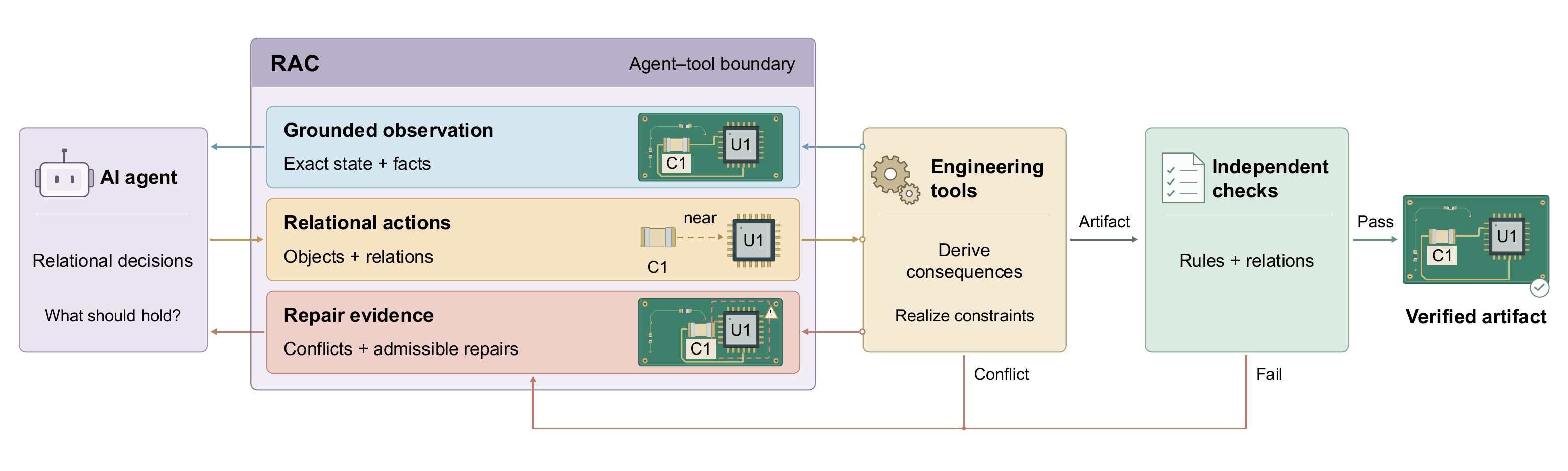}
\caption{Overview of \system. \system defines the agent--tool boundary. The agent observes the exact engineering state. Through \emph{relational actions}, the agent states relations among named objects, for example placing capacitor C1 \emph{near} U1, without writing coordinates. Engineering tools derive the consequences of these relations and realize them jointly as a concrete artifact. Independent checks then evaluate the artifact against both design rules and the stated relations. A passing artifact is returned as verified. A realization conflict or a failed check returns \emph{repair evidence}, which names the conflicting relations and the admissible repairs, and the agent uses it to revise its relational decision.}
\label{fig:responsibility-boundary}
\end{figure*}

In this paper, we develop \system, a Relational Agent Contract, to define responsibility boundaries.
\system is \emph{harness-agnostic}: it specifies the tool contract a harness exposes. 
Within that harness, \system defines authoritative observations, semantic actions, tool-based realization, independent artifact verification, and feedback for repair. 
It requires no model retraining. 

This boundary separates observation from construction. The agent may inspect exact coordinates while expressing only relations such as ``C1 close to U1.'' Tools jointly realize committed relations; independent checks evaluate both engineering rules and the required relations on the resulting board. Failed checks return evidence for the next relational decision. Thus, \system removes judgment--derivation coupling while retaining explicit semantic control.

We test \system in across constraint-governed engineering tasks from three different domains. 
With the harness held fixed, \system achieves the highest clean-pass rate for all nine models on PCB tasks. In scheduling tasks, \system improves the final quality-pass rate for seven models and ties with a baseline at 100\% for the other two.
Moreover, \system consumes less tokens and incurs less wall time latency for most of evaluated models.

\section{Related Work}
\label{sec:related}

\paragraph{Agent interfaces and protocols.}
Language model tool use has been established by \citet{yao2023react,schick2023toolformer}, while \citet{sweagent2024,wang2024codeact,lu2024axis} show that action and observation design materially affects agent performance. 
Richer interface wrappers can also repair mismatches between an agent's interpretation and environment logic without changing either component~\citep{liu2025agentenvalign}, and benchmarks judge agents by the resulting artifacts~\citep{jimenez2024swe,merrill2026terminalbench,Terminal-Bench-Science_2026}. 
Model context protocol (MCP) \citep{mcp2026} standardizes tool discovery and invocation. 
Anthropic recently released model hardware standard (MHS), extending discoverable access to physical equipment~\citep{anthropic2026mhs}.
Surveys of agent protocols can be found in \citep{ehtesham2025protocols}, along with threat analysis \citep{anbiaee2026threat}. 
These systems make capabilities usable and connectable, but do not determine the semantics or verifiability of the actions they carry. 
\system is complementary: its contract specifies the domain semantics, derivation boundary, independent verification, and repair evidence that can travel through such connections.

\paragraph{Language-model agents for hardware design.}
Surveys of language models for EDA identify compiler- and simulator-driven repair as a dominant pattern and classify systems by the handoff between agent and tool~\citep{pan2025llm4eda,liu2026agenticeda}. At the board level, HWE-Bench \citep{qiu2026hwebench} and PCBSchemaGen \citep{zou2026pcbschemagen} expose pin-level failures and the value of deterministic verification. 
PCBAgent \citep{ChenPCBAgent2025} and LayoutCopilot \citep{liu2025layoutcopilot} combine language-model interaction with learned or conventional layout tools; 
PCBWorld grounds agents in KiCad operations and design-rule feedback~\citep{song2026pcbworld}; and OmniLayout \citep{lu2026omnilayout} and OmniRouting \citep{lu2026omnirouting} document the difficulty of directly emitting placements, tracks, vias, and layers that satisfy design. 
Embedded-agent systems similarly improve generation through retrieval, compilation, execution, and physical-device feedback, while exposing implicit platform and safety obligations~\citep{xu2025embedagent,embedeval2026,garcia2026closedloop,zhang2026embedded}. 
Across these systems, tool grounding does not itself specify which concrete consequences should remain model-authored.

\paragraph{Verification and repair.}
Self-Debugging \citep{chen2024selfdebug}, Reflexion \citep{shinn2023reflexion}, and Self-Refine \citep{madaan2023selfrefine} use execution results or critiques to revise outputs, with the strongest gains arising from independent test oracles. 
Formal systems tighten this loop: Clover \citep{sun2024clover} verifies generated programs with Dafny \citep{leino2023program}, while AlphaVerus \citep{aggarwal2025alphaverus} also shows that a model can satisfy a verifier vacuously without an independent alignment check.
Program-based verifier ensembles likewise improve spatial generation, and EDA repair loops become less effective as rule complexity grows~\citep{zhang2026weakverifiers,liu2026agenticeda}.
\system therefore uses checkers independent of artifact production and returns semantic repair evidence, so repair operates on intent instead of geometry or register values.

\paragraph{Semantic intermediates and deterministic realization.}
Models remain weak at coordinate-based geometric computation~\citep{truong2026gpsbench,blink2024}.
Several systems delegate formal consequences to conventional algorithms: LLM+P translates language into PDDL \citep{liu2023llmp}, Code as Policies \citep{liang2023codeaspolicies} emits programs over grounded APIs, and SayPlan \citep{sayplan2023} plans over semantic scene hierarchies. 
Holodeck \citep{holodeck2024} compiles relational scene constraints and finds that direct coordinate emission performs no better than random placement, whereas LayoutGPT \citep{feng2023layoutgpt} emits coordinates directly. 
Imperative relative-placement programs can outperform declarative constraint formulations~\citep{gumin2025imperative}, formal intermediate languages produce large accuracy differences in logical reasoning~\citep{beiser2025intermediate}, and geometric invalidity rises sharply with CAD topology complexity~\citep{wang2026text2cadbench}. 
FunSearch \citep{romeraparedes2024funsearch}, AlphaEvolve \citep{novikov2025alphaevolve}, and AlphaGeometry \citep{alphageometry2024} similarly pair learned proposals with deterministic evaluation or deduction; in electronic design, ALIGN compiles topology and relational constraints into analog layouts~\citep{align2021}. 
This literature establishes semantic-to-formal compilation and shows that the model-facing representation governs outcomes. 
\system makes this division explicit and testable.

\section{Constraint-Governed Engineering Tasks}\label{sec:framework}

We describe an engineering task instance by
\begin{equation} \label{eq:task-instance}
    \tau=(c,r,\mathcal{Y},\Phi),
\end{equation}
where $c$ is the engineering context including the authoritative facts fixed for this instance, such as a PCB design or a particular microcontroller configuration. 
The requirement $r$ states what the artifact should accomplish. 
The set $\mathcal{Y}$ is the artifact space in which a solution may be constructed (e.g., placed-and-routed boards or embedded programs). 
Finally, $\Phi=\{\phi_1,\ldots,\phi_m\}$ is the set of validity constraints for the instance. Each constraint is a checkable predicate $\phi_i(c,r,y)\in\{0,1\}.$
An artifact $y$ is considered a valid solution to a constraint-governed engineering task if every constraint in $\Phi$ holds.

Our formulation in \eqref{eq:task-instance} covers a wide range of typical situations.
A \emph{construction task} asks for creating any valid artifact $y$. 
A \emph{repair task} additionally provides an initial artifact $y_0$ within context $c$ and a preservation requirement included among the predicates in $\Phi$.

\begin{figure*}[t]
\vspace{-2em}
    \centering
    \includegraphics[width=\textwidth]
        {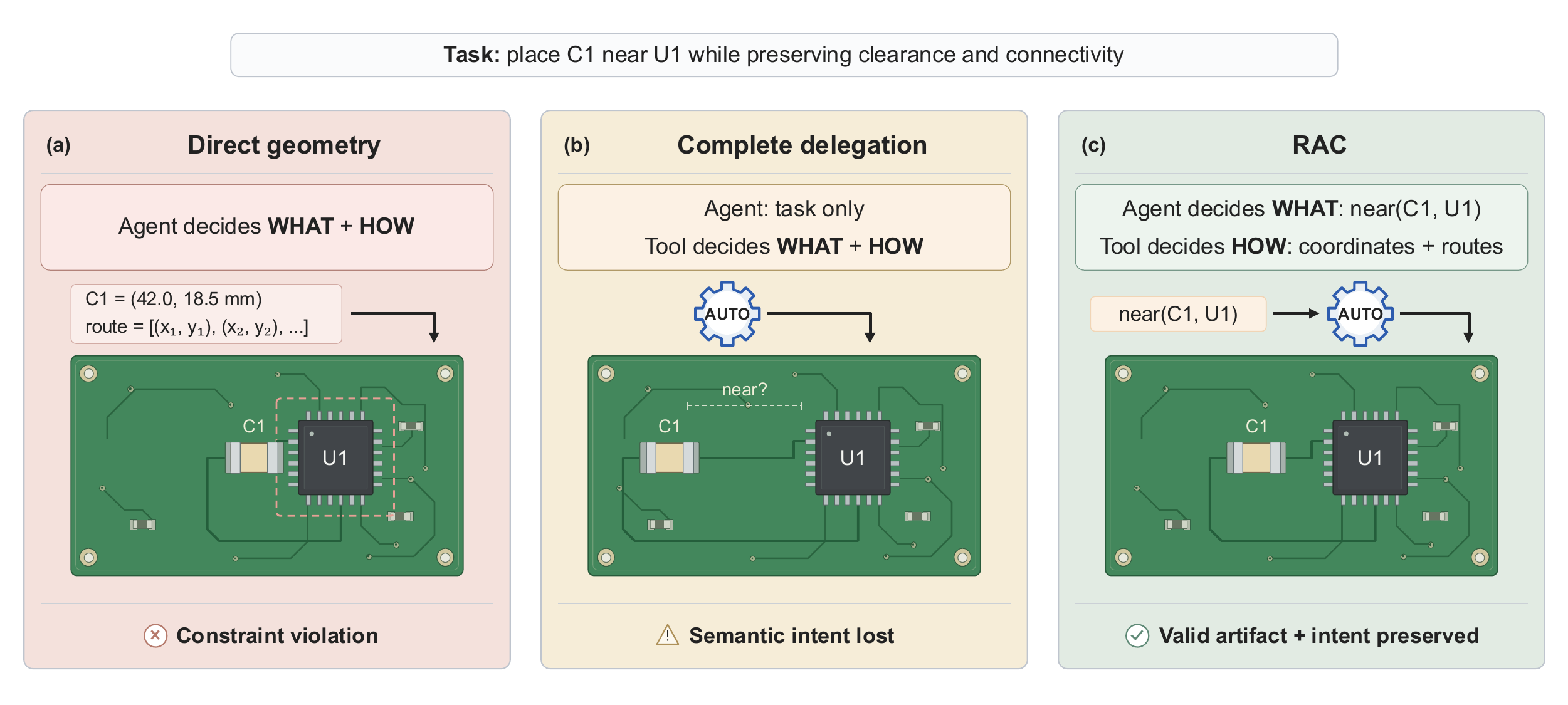}
    \vspace{-2em}
    \caption{
    Illustration of responsibility allocation in a
    constraint-governed PCB task.
    Existing geometry methods (left) directly assign both semantic decisions and metric realization to the agent, exposing the artifact to coordinate-level constraint violations.
    Complete delegation of both decisions to engineering tools (middle) can produce valid geometry but lose the intended relation.
    \system (right) preserves the semantic decision with the agent (e.g., placing C1 near U1) while engineering tools derive the coordinates and routes. 
    This division preserves intent while enabling deterministic realization and verification.
    }
    \label{fig:baseline compare}
\end{figure*}

\paragraph{PCB instantiation.}
For PCB layout, context $c$ contains the board outline, footprints and pads, netlist, layer stack, existing copper, and design rules. The requirement $r$ specifies placement, routing, or repair intent; $\mathcal{Y}$ contains candidate board layouts; and $y$ records component placement and copper geometry. The constraints $\Phi$ include geometric legality, clearance, connectivity, and task-specific relations such as proximity, alignment, avoidance, differential pairing, and length matching. These predicates share component positions and copper paths. Moving one component can simultaneously alter clearance, routability, and relation satisfaction, making feasibility intrinsically coupled.

\paragraph{Embedded instantiation.}
For embedded development, $c$ contains the selected microcontroller and board, devicetrees, register descriptions, build configuration, architecture facts, and errata. The requirement $r$ specifies the desired program behavior; $\mathcal{Y}$ contains candidate programs; and $y$ is the program source and resulting build artifact. The constraints $\Phi$ combine functional correctness with context-specific obligations for memory reachability, cache maintenance, alignment, interrupt use, clocking, pins, power domains, and errata. 
The same source-level requirement can therefore admit different artifacts on different hardware. 
Moreover, one implementation choice (e.g., placing a direct memory access (DMA) buffer in cached memory) can activate several obligations at once.

\section{\system: A Relational Agent Contract}\label{sec:RAP}

\paragraph{Design principle.}

\system separates responsibilities between agent and engineering tools. 
It lets agents retain decisions that require semantic engineering judgment, while tools derive consequences that follow mechanically from the engineering context and those decisions.
That is, interpreting the requirement $r$ remains the agent's responsibility; computing what an accepted decision implies for the particular context $c$ and current artifact $y$ belongs to the engineering tools.

\paragraph{Contract design.}
\system splits responsibilities between agent and domain tools using an interface contract $\mathcal{C}=(\mathcal{O},\mathcal{A},E,V,F)$. 
The agent makes observations following a typed schema $\mathcal{O}$, populated by the domain tool from the authoritative context $c$, the requirement $r$, and the current artifact $y$. 
The agent then takes action $a\in\mathcal{A}$, comprising a set of \emph{commitments}: semantic choices that add, revise, or remove relations the artifact should satisfy, such as placing C1 near U1. $\mathcal{A}$ omits every field that tools can derive from context $c$.
Domain tools derive what the committed relations imply in context $c$ and realize all of them together in the next artifact $y'=E(c,y,a)$. 
Here, $E$ is deterministic or seed-controlled derivation, so every realization can be replayed. It also reports which relations were realized on $y'$ and which were not.
An independent checker $V$ evaluates every predicate in $\Phi$ on $(c,r,y')$. 
Feedback $F$ turns derivation's execution report and independent checker's per-predicate verdicts into evidence about artifact, for example which clearance check failed and between which objects. The harness delivers this evidence to the agent together with the next observation, and the agent uses it to revise its commitments.

\section{Evaluations of \system with PCB Layout Synthesis}
\label{sec:setup}

\subsection{Experimental Setup}

\paragraph{Models.}
We evaluate nine models covering both open-source and closed-source ones: DeepSeek-V4-Flash \citep{deepseekai2026deepseekv4}, GLM-5.3-Flash \citep{glm5team2026glm5vibecodingagentic}, GPT-5.6-Luna, GPT-5.6-Sol \citep{openai2026gpt56}, Kimi-K2.7 \citep{kimi2026k27}, Qwen3.6-27B \citep{qwen3.6-27b}, Claude-Opus-4.8 \citep{anthropic2026claudeopus48}, Claude-Opus-5 \citep{anthropic2026claudeopus5}, and Gemini-3.8-Flash \citep{google2026gemini}.
We use medium reasoning effort for GPT-5.6 Sol and disable reasoning for DeepSeek V4 Flash.
All other models use default reasoning effort and temperature.

\paragraph{Tasks.}
\system is evaluated on 98 tasks.
The task suite comprises routing (42), placement (31), combined placement-and-routing (17), and repair (8) tasks. 
These tasks span decoupling, board-edge association, keepouts, grouping, symmetry, alignment, and routing relations. 

\paragraph{Metrics.}
We evaluate each task using its generated artifact and report the following metrics:
\begin{itemize}
       \item \textbf{Clean-pass rate.} An artifact $y_\tau$ from a task $\tau$ is considered clean pass if $\bigwedge_{\phi\in\Phi_\tau}\phi(c,r,y_\tau)=1$, where $\Phi_\tau$ is the set of constraints associated with task $\tau$. Clean-pass rate is defined as 
        \begin{equation*}
            CPR=\frac{1}{|\mathcal{T}|}\left |
            \left\{\tau\in \mathcal{T}: \bigwedge_{\phi\in\Phi_\tau}\phi(c,r,y_\tau)=1\right\}\right |.
        \end{equation*}
       \item \textbf{Computational cost.}
       We report the input and output token counts, and the wall-clock time from the start of execution to termination.
   \end{itemize}

\paragraph{Baselines and controls.}
We compare \system with baseline interfaces below that assign geometric and semantic responsibility differently. An illustrative comparison between \system and baselines is presented in Figure \ref{fig:baseline compare}.
\begin{itemize}
\item \textbf{Coord. (Direct coordinate generation):} the agent specifies component positions and routing paths directly, following the geometric output interfaces evaluated by PCBWorld, OmniLayout, and OmniRouting~\citep{song2026pcbworld,lu2026omnilayout,lu2026omnirouting}. This baseline measures performance when the agent owns geometric realization.
\item \textbf{Coord.+Py (Arithmetic-assisted coordinate generation):} the agent retains the same coordinate-level interface as \textbf{Coord.} but receives a Python sandbox for geometric calculations. This variant tests whether access to computation, rather than the action representation, explains the difficulty of direct geometry generation.
\item \textbf{Candidates (Tool-generated candidate selection):} the agent selects among geometric proposals generated by the toolchain, adapting PCBWorld's engine-grounded interaction model~\citep{song2026pcbworld}. This baseline measures the benefit of solver-generated proposals without introducing a relational write interface.
\item \textbf{Seq. Relations (Sequential relational actions):} the agent submits one semantic relation at a time, following the imperative organization studied for solver-backed scene generation~\citep{gumin2025imperative}.
\item \textbf{Req.-Only (Requirements-only delegation):} the agent supplies high-level requirements and delegates placement and routing decisions to the toolchain, reflecting the broader division of labor used by automated layout systems~\citep{align2021,ChenPCBAgent2025}.
\end{itemize}

\paragraph{Harness.}
All interfaces and baselines run in a shared harness, which we call the reference harness; Section~\ref{sec:harness} replaces it with other agent harnesses.
At each turn, the harness presents the typed board state, requests an action in the interface's schema, executes it with the shared placement and routing backend, and returns the resulting board state and check results.
An episode ends when the agent finishes or after at most five model calls.

\begin{table*}[t]
    \centering
    \vspace{-1em}
    \caption{Clean-pass rates (\%) on 98 tasks across six random seeds. The final row averages the nine model-specific rates. Bold marks the highest clean-pass rate in each row. \system consistently outperforms all baselines across all agents.}
    \label{tab:construction-main}
    \setlength{\tabcolsep}{2.2pt}
    \renewcommand{\arraystretch}{1.14}
    \newcommand{\passcell}[1]{#1\%}
    \begin{tabular}{@{}lcccccc@{}}
        \toprule
        Model & Coord. & Coord.+Py & Candidates & \shortstack{Seq.\\Relations} & Req.-Only & \textbf{\system} \\
        \midrule
        \leaderlogo{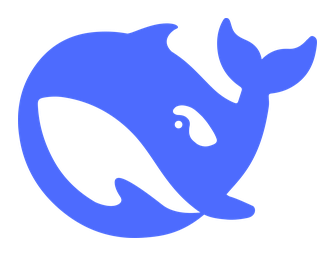}\ DeepSeek-V4-Flash
          & \passcell{10.2}
          & \passcell{7.1}
          & \passcell{37.8}
          & \passcell{64.3}
          & \passcell{74.5}
          & \passcell{\textbf{92.9}} \\
        \leaderlogo{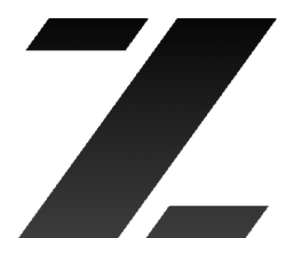}\ GLM-5.3-Flash
          & \passcell{25.5}
          & \passcell{20.4}
          & \passcell{33.7}
          & \passcell{60.2}
          & \passcell{68.4}
          & \passcell{\textbf{89.8}} \\
        \leaderlogo{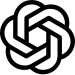}\ GPT-5.6-Luna
          & \passcell{24.5}
          & \passcell{26.5}
          & \passcell{41.8}
          & \passcell{73.5}
          & \passcell{77.6}
          & \passcell{\textbf{95.9}} \\
        \leaderlogo{assets/logos/openai_logo.png}\ GPT-5.6-Sol
          & \passcell{29.6}
          & \passcell{26.5}
          & \passcell{40.8}
          & \passcell{65.3}
          & \passcell{77.6}
          & \passcell{\textbf{92.9}} \\
        \leaderlogo{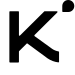}\ Kimi-K2.7
          & \passcell{6.1}
          & \passcell{4.1}
          & \passcell{22.4}
          & \passcell{43.9}
          & \passcell{61.2}
          & \passcell{\textbf{77.6}} \\
        \leaderlogo{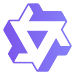}\ Qwen3.6-27B
          & \passcell{26.5}
          & \passcell{25.5}
          & \passcell{34.7}
          & \passcell{62.2}
          & \passcell{77.6}
          & \passcell{\textbf{93.9}} \\
        \leaderlogo{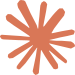}\ Claude-Opus-4.8
          & \passcell{25.5}
          & \passcell{24.5}
          & \passcell{43.9}
          & \passcell{60.2}
          & \passcell{74.5}
          & \passcell{\textbf{91.8}} \\
        \leaderlogo{assets/logos/claude_flower.png}\ Claude-Opus-5
          & \passcell{29.6}
          & \passcell{24.5}
          & \passcell{39.8}
          & \passcell{61.2}
          & \passcell{74.5}
          & \passcell{\textbf{89.8}} \\
        \leaderlogo{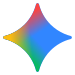}\ Gemini-3.8-Flash
          & \passcell{31.6}
          & \passcell{29.6}
          & \passcell{41.8}
          & \passcell{70.4}
          & \passcell{80.6}
          & \passcell{\textbf{93.9}} \\
        \midrule
        \textbf{Average}
          & \passcell{23.2}
          & \passcell{21.0}
          & \passcell{37.4}
          & \passcell{62.4}
          & \passcell{74.0}
          & \passcell{\textbf{90.9}} \\
        \bottomrule
    \end{tabular}
\end{table*}

\paragraph{Instantiating \system for PCB tasks.}
We instantiate the contract of Section~\ref{sec:RAP} for PCB tasks.
The observation schema $\mathcal{O}$ is a typed view of the authoritative board state, including components, pins, nets, existing copper, unrouted connections, and design-rule results.
Exact geometry remains observable because the agent needs accurate evidence to make semantic decisions.
The responsibility boundary is enforced in $\mathcal{A}$, which permits only relations over named objects, such as proximity, alignment, symmetry, edge association, keepouts, matched routing, layer preference, and via limits.
No action accepts a coordinate or a metric distance: magnitudes use small ordered classes (for example, \code{tight}, \code{close}, or \code{loose} placement distance), and their metric values are fixed on the solver side.
This preserves the agent's choice of what the layout should satisfy while assigning mechanically derivable coordinates, paths, and clearances to $E$, implemented by the shared, seed-controlled placement and routing backend.
Because these relations are coupled, \system submits a complete set of committed relations as one bundle that $E$ realizes jointly.
The resulting artifact is evaluated by $V$ using an independent design-rule checker and relation-specific geometric oracles that do not consume the solver's own satisfaction labels.
Feedback $F$ returns the realized state and verification evidence for the next turn.
When realization is infeasible, this evidence identifies the failure type, conflicting constraints, violation magnitudes, and admissible relaxations, so the agent can revise its commitments instead of repairing geometry.

\begin{figure*}[t]
    \centering
    \includegraphics[width=0.86\textwidth]{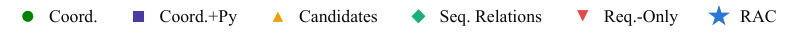}\\[0.3em]
    \begin{tabular}{@{}c@{\hspace{0.4em}}c@{}}
        \rotatebox[origin=c]{90}{\small Clean-pass rate (\%)} &
        \begin{minipage}{0.955\textwidth}
            \centering
            \includegraphics[width=0.328\linewidth]{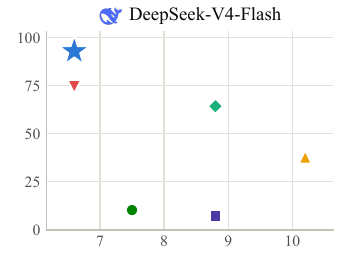}\hfill
            \includegraphics[width=0.328\linewidth]{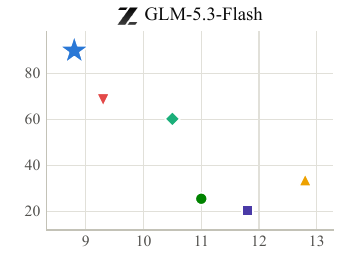}\hfill
            \includegraphics[width=0.328\linewidth]{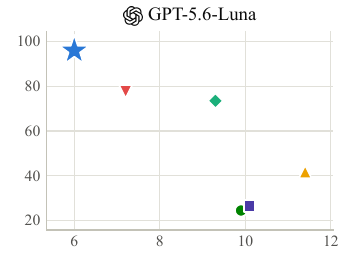}\\[0.2em]
            \includegraphics[width=0.328\linewidth]{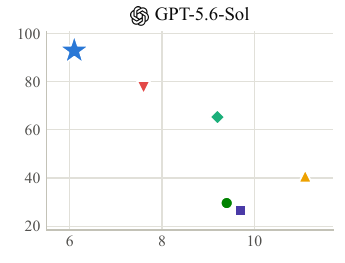}\hfill
            \includegraphics[width=0.328\linewidth]{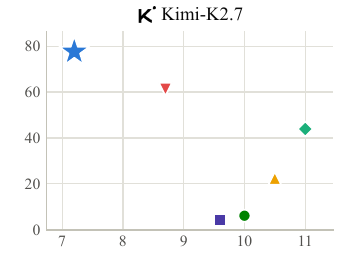}\hfill
            \includegraphics[width=0.328\linewidth]{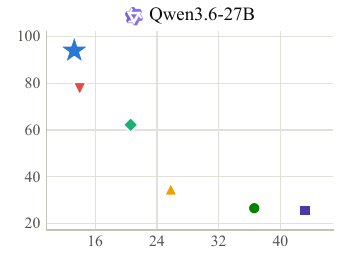}\\[0.2em]
            \includegraphics[width=0.328\linewidth]{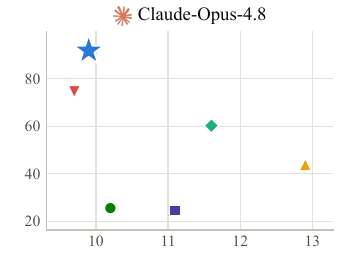}\hfill
            \includegraphics[width=0.328\linewidth]{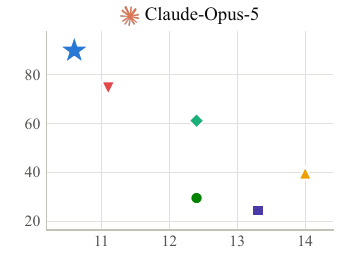}\hfill
            \includegraphics[width=0.328\linewidth]{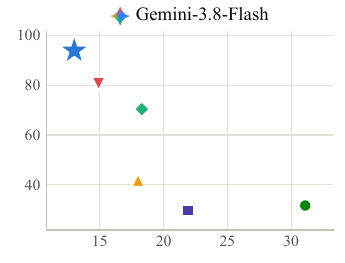}
        \end{minipage} \\[0.2em]
        & {\small Tokens per episode (thousands)}
    \end{tabular}
    \vspace{-0.5em}
    \caption{Clean-pass rate against average token use per episode on the PCB tasks, one panel per model. The token axis is logarithmic. Higher and further left is better. \system attains the highest clean-pass rate for every model and the lowest token use for eight of the nine, tying Req.-Only on DeepSeek-V4-Flash.}
    \label{fig:pcb-pareto}
\end{figure*}

\subsection{Experimental Results}

\paragraph{\system consistently improves clean-pass rate.}
Table~\ref{tab:construction-main} compares \system against all baselines on all evaluation tasks. 
We observe that \system consistently achieves the highest clean-pass rate across all models, ranging from 77.6\% to 95.9\%.
Compared to tool-generated candidate selection, \system improves the mean clean-pass rate by 53.5 percentage points. 

\begin{table*}[t]
    \centering
    \caption{Average wall-clock latency (seconds) per episode on PCB tasks. \system's wall time is comparable within a model under the matched transport and worker regime.}
    \label{tab:construction-latency}
    \setlength{\tabcolsep}{4.5pt}
    \renewcommand{\arraystretch}{1.12}
    \begin{tabular}{@{}lcccccc@{}}
        \toprule
        Model & Coord. & Coord.+Py & Candidates & \shortstack{Seq.\\Relations} & Req.-Only & \textbf{\system} \\
        \midrule
        \leaderlogo{assets/logos/Deepseek.png}\ DeepSeek-V4-Flash
          & 12.5 & 10.3 & 13.0 & 11.2 & 9.2 & \textbf{9.0} \\
        \leaderlogo{assets/logos/glm.png}\ GLM-5.3-Flash
          & 38.2 & 41.7 & 35.2 & 20.4 & 22.4 & \textbf{19.4} \\
        \leaderlogo{assets/logos/openai_logo.png}\ GPT-5.6-Luna
          & 41.8 & 66.9 & 23.4 & 14.9 & 14.8 & \textbf{9.1} \\
        \leaderlogo{assets/logos/openai_logo.png}\ GPT-5.6-Sol
          & 59.1 & 62.0 & 37.6 & 21.7 & 25.5 & \textbf{19.5} \\
        \leaderlogo{assets/logos/moonshot.png}\ Kimi-K2.7
          & 29.4 & 37.7 & 20.7 & 22.9 & 21.0 & \textbf{13.9} \\
        \leaderlogo{assets/logos/qwen.png}\ Qwen3.6-27B
          & 409.4 & 511.1 & 194.1 & 157.4 & 109.8 & \textbf{88.8} \\
        \leaderlogo{assets/logos/claude_flower.png}\ Claude-Opus-4.8
          & 20.4 & 31.8 & 16.2 & \textbf{11.4} & 16.1 & 14.0 \\
        \leaderlogo{assets/logos/claude_flower.png}\ Claude-Opus-5
          & 59.0 & 71.7 & 35.0 & \textbf{22.4} & 24.9 & 22.8 \\
        \leaderlogo{assets/logos/gemini.png}\ Gemini-3.8-Flash
          & 162.7 & 107.4 & 65.5 & 67.2 & 61.2 & \textbf{49.3} \\
        \bottomrule
    \end{tabular}
\end{table*}

\paragraph{\system identifies the effective responsibility boundary.}
The average row of Table~\ref{tab:construction-main} shows how clean-pass rate changes as geometric and semantic responsibilities are reassigned between the agent and the toolchain. Direct coordinate generation reaches 23.2\% on average, and adding Python-based arithmetic assistance does not improve it (21.0\%). Letting the toolchain generate geometric candidates raises the rate to 37.4\%, while allowing the agent to express semantic relations sequentially raises it further to 62.4\%. \system reaches 90.9\% by preserving the agent's control over the complete set of relations while jointly realizing its coupled geometry. Delegating the semantic decisions as well, as in requirements-only delegation, only reaches 74.0\%.

\paragraph{The engineering contract matters more than model selection.}
With \system, eight of the nine models reach 89.8--95.9\%, exceeding the best result obtained by any model with any baseline.
The exception, Kimi-K2.7, reaches 77.6\% with \system, still 16.3 percentage points above its own best baseline.
This result shows that model capability alone cannot compensate for an interface that assigns geometric and semantic responsibilities poorly.

\paragraph{\system improves token efficiency as well as accuracy.} 
Figure \ref{fig:pcb-pareto} and Table~\ref{tab:construction-tokens} show that \system has the lowest average token use for eight of the nine models.
Relative to tool-generated candidate selection, it reduces token use for all nine models by 22.9--48.5\%, while simultaneously improving mean clean-pass rate from 37.4\% to 90.9\%.
This efficiency arises because the agent can express the complete set of relations in a compact bundle, rather than repeatedly inspecting and selecting among geometric proposals or constructing coordinates over multiple turns. Requirements-only delegation can use comparably few tokens, but its clean-pass rates are substantially lower than those of \system.

\paragraph{\system's accuracy gains do not come at the cost of greater latency.} Within every model, \system is faster than tool-generated candidate selection, reducing mean wall-clock time by 13.6--61.1\%, and it achieves the lowest latency among all interfaces for seven of the nine models (Table \ref{tab:construction-latency}). Candidate selection incurs substantial overhead by repeatedly generating, presenting, and evaluating geometric alternatives, whereas \system communicates the complete set of relations and realizes the coupled constraints jointly. 
These results show that \system improves engineering reliability while shortening the interaction required to produce a verified artifact.
Hence, \system provides a quality–efficiency improvement, not a compute trade-off.

\paragraph{Engineering performance is not monotonic in nominal model capability.} 
Table~\ref{tab:construction-main} shows that newer or higher-tier models do not consistently achieve higher clean-pass rates. 
Under \system, GPT-5.6-Luna exceeds GPT-5.6-Sol (95.9\% vs. 92.9\%), and Claude-Opus-4.8 exceeds Claude-Opus-5 (91.8\% vs. 89.8\%).
In contrast, changing the agent--tool interface produces a large and consistent improvement across all nine models. 
This indicates that success on engineering tasks depends not only on the underlying model, but also on whether the interface assigns geometric realization and semantic decision-making to the components best suited to them.
These results suggest that selecting a nominally stronger model is not a substitute for designing the right engineering contract.

\subsection{Robustness across Agent Harnesses}
\label{sec:harness}

\paragraph{Setup.}
We evaluate GPT-5.6-Luna and GPT-5.6-Sol on the same 98 tasks and compare \system with \textbf{Coord.}
Both interfaces are exposed as tools.
Four harnesses provide these tools to the model.
\textbf{Native} is a minimal loop that offers them through the model's native function-calling interface.
The other three run in Harbor~0.23.0 \citep{Harbor_Framework}, and use one of three coding agents: \textbf{Codex}~0.157.1, \textbf{Claude Code}~2.1.294, or \textbf{OpenCode}~1.18.34.
These agents reach the same tools through the Model Context Protocol~\citep{mcp2026}, with web search and sub-agents disabled.
In every harness, model calls pass through one metering bridge that counts each call against a budget of five model calls, and applies medium reasoning effort.

\begin{table*}[t]
    \centering
    \caption{Clean-pass rates and average token use per task of \textbf{Coord.} and \system on the 98 PCB test tasks under four agent harnesses. $\Delta$ is \system minus Coord.; $\uparrow$/$\downarrow$ mark whether higher or lower is better. Within every harness and model, \system achieves a higher clean-pass rate with fewer tokens.}
    \label{tab:harness}
    \setlength{\tabcolsep}{3.0pt}
    \renewcommand{\arraystretch}{1.12}
    \begin{tabular}{@{}lcccccccccccc@{}}
        \toprule
         & \multicolumn{6}{c}{\leaderlogo{assets/logos/openai_logo.png}\ GPT-5.6-Luna} & \multicolumn{6}{c}{\leaderlogo{assets/logos/openai_logo.png}\ GPT-5.6-Sol} \\
        \cmidrule(lr){2-7}\cmidrule(lr){8-13}
        & \multicolumn{3}{c}{Clean pass $\uparrow$} & \multicolumn{3}{c}{Tokens / task $\downarrow$}
        & \multicolumn{3}{c}{Clean pass $\uparrow$} & \multicolumn{3}{c}{Tokens / task $\downarrow$} \\
        \cmidrule(lr){2-4}\cmidrule(lr){5-7}\cmidrule(lr){8-10}\cmidrule(lr){11-13}
        Harness & Coord. & \system & $\Delta$ & Coord. & \system & $\Delta$ & Coord. & \system & $\Delta$ & Coord. & \system & $\Delta$ \\
        \midrule
          \phantom{\leaderlogo{assets/logos/openai_logo.png}}\ Native
          & 29.6\% & \textbf{92.9\%} & \diffcell{$+$63.3} & 22.9k & \textbf{16.1k} & \diffcellneg{$-$6.8k}
          & 26.5\% & \textbf{95.9\%} & \diffcell{$+$69.4} & 22.4k & \textbf{14.5k} & \diffcellneg{$-$7.9k} \\
        \leaderlogo{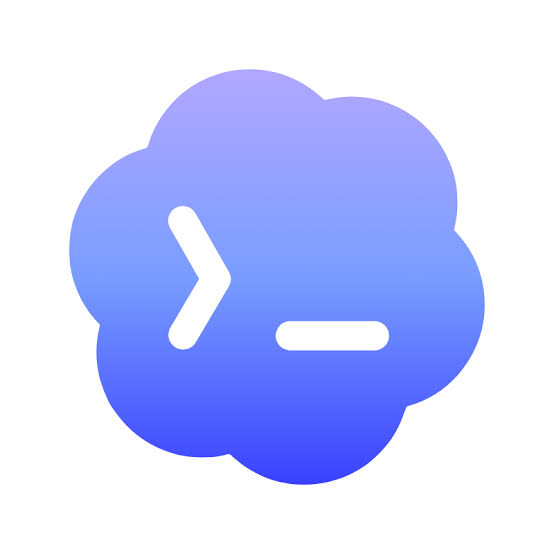}\ Codex
          & 24.5\% & \textbf{91.8\%} & \diffcell{$+$67.3} & 71.4k & \textbf{53.2k} & \diffcellneg{$-$18.2k}
          & 22.7\% & \textbf{93.9\%} & \diffcell{$+$71.2} & 69.8k & \textbf{64.9k} & \diffcellneg{$-$4.9k} \\
        \leaderlogo{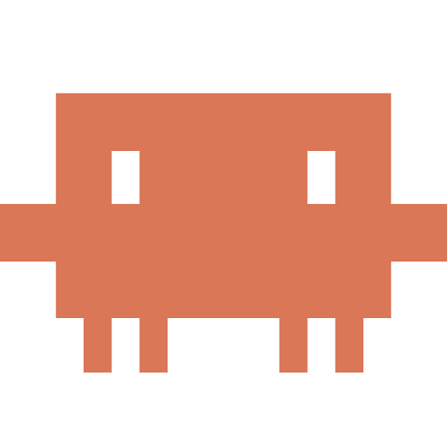}\ Claude Code
          & 26.5\% & \textbf{91.8\%} & \diffcell{$+$65.3} & 81.9k & \textbf{60.0k} & \diffcellneg{$-$21.9k}
          & 32.7\% & \textbf{94.9\%} & \diffcell{$+$62.2} & 84.8k & \textbf{78.8k} & \diffcellneg{$-$6.0k} \\
        \leaderlogo{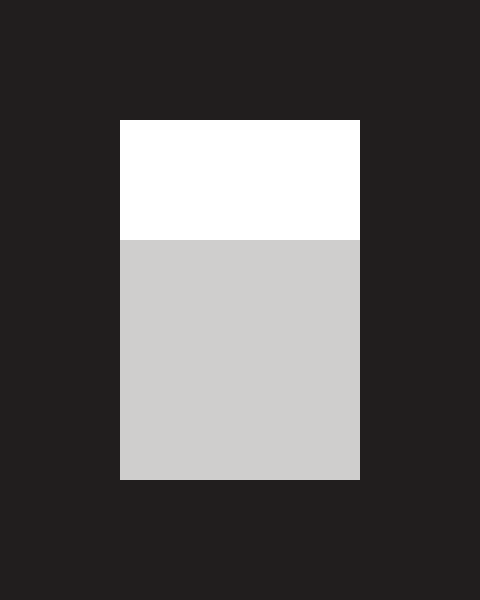}\ OpenCode
          & 25.5\% & \textbf{92.9\%} & \diffcell{$+$67.4} & 47.4k & \textbf{35.1k} & \diffcellneg{$-$12.3k}
          & 26.5\% & \textbf{93.9\%} & \diffcell{$+$67.4} & 47.6k & \textbf{38.4k} & \diffcellneg{$-$9.2k} \\
         \bottomrule
    \end{tabular}
\end{table*}

\paragraph{The gain from \system holds in every harness.}
Table~\ref{tab:harness} shows that, within each harness and model, \system raises clean pass from 22.7--32.7\% to 91.8--95.9\%.
The gain of \system persists across different harnesses.
Under \system, GPT-5.6-Luna passes 90--91 tasks and GPT-5.6-Sol 92--94 tasks across the four harnesses.

\paragraph{Harnesses differ mainly in cost.}
With \system, the three Harbor agents use 2.2--5.4 times the tokens of Native per task, mostly as input tokens that each call spends on the agent's own instructions and tool list.
Within every harness, \system uses fewer tokens per task than \textbf{Coord.}
The choice of harness therefore changes the cost of using \system on PCB tasks but not the gain it delivers.

\paragraph{Additional Results.}
Additional experimental results on PCB tasks along with examples are in Appendix \ref{app:additional-pcb-results}.
Evaluations using embedded development are in Appendix \ref{app:embedded-results}.

\section{Evaluations of \system with Job Shop Scheduling}
\label{sec:scheduling-main}

\subsection{Experimental Setup}

\paragraph{Agents and tasks.}
We evaluate on the 30 public Lawrence job-shop instances LA01--LA30 from JSPLIB \citep{JSPLib}.
These instances span 10--20 jobs, five or ten machines, and 50--200 operations.
Each job specifies an ordered sequence of operations with fixed machines and processing times. 
A valid schedule must preserve job precedence and prevent machine overlap.
Each model completes one episode per instance and interface, giving $9\times30\times2=540$ episodes and 270 matched model--instance pairs.
Every episode contains three successive full proposals, and the final proposal is scored.

\begin{table*}[t]
       \centering
       \caption{Final proposal and first proposal quality-pass rates (\%) on LA01--LA30. $\Delta$ is the gain of \system over Code in percentage points. \system improves scheduling action delivery and delivers usable schedules earlier.}
       \label{tab:scheduling-main-quality}
       \setlength{\tabcolsep}{7.0pt}
       \renewcommand{\arraystretch}{1.12}
       \begin{tabular}{@{}lcccccc@{}}
           \toprule
           & \multicolumn{3}{c}{Final proposal}
           & \multicolumn{3}{c}{First proposal} \\
           \cmidrule(lr){2-4}\cmidrule(l){5-7}
           Model & Code & \textbf{\system} & $\Delta$ & Code & \textbf{\system} & $\Delta$ \\
           \midrule
           \leaderlogo{assets/logos/Deepseek.png}\ DeepSeek-V4-Flash
             & 26.7\% & \textbf{100.0}\% & \diffcell{+73.3}
             & 10.0\% & \textbf{100.0}\% & \diffcell{+90.0} \\
           \leaderlogo{assets/logos/glm.png}\ GLM-5.3-Flash
             & 63.3\% & \textbf{100.0}\% & \diffcell{+36.7}
             & 46.7\% & \textbf{100.0}\% & \diffcell{+53.3} \\
           \leaderlogo{assets/logos/openai_logo.png}\ GPT-5.6-Luna
             & 86.7\% & \textbf{100.0}\% & \diffcell{+13.4}
             & 83.3\% & \textbf{96.7}\% & \diffcell{+13.3} \\
           \leaderlogo{assets/logos/openai_logo.png}\ GPT-5.6-Sol
             & \textbf{100.0}\% & \textbf{100.0}\% & \diffcell{0.0}
             & \textbf{100.0}\% & \textbf{100.0}\% & \diffcell{0.0} \\
           \leaderlogo{assets/logos/moonshot.png}\ Kimi-K2.7
             & 30.0\% & \textbf{96.7}\% & \diffcell{+66.7}
             & 23.3\% & \textbf{100.0}\% & \diffcell{+76.7} \\
           \leaderlogo{assets/logos/qwen.png}\ Qwen3.6-27B
             & 33.3\% & \textbf{70.0}\% & \diffcell{+36.7}
             & 20.0\% & \textbf{46.7}\% & \diffcell{+26.7} \\
           \leaderlogo{assets/logos/claude_flower.png}\ Claude-Opus-4.8
             & 16.7\% & \textbf{100.0}\% & \diffcell{+83.3}
             & 6.7\% & \textbf{100.0}\% & \diffcell{+93.3} \\
           \leaderlogo{assets/logos/claude_flower.png}\ Claude-Opus-5
             & \textbf{100.0}\% & \textbf{100.0}\% & \diffcell{0.0}
             & 26.7\% & \textbf{100.0}\% & \diffcell{+73.3} \\
           \leaderlogo{assets/logos/gemini.png}\ Gemini-3.8-Flash
             & 80.0\% & \textbf{100.0}\% & \diffcell{+20.0}
             & 60.0\% & \textbf{100.0}\% & \diffcell{+40.0} \\
           \midrule
           \textbf{Average}
             & 59.6\% & \textbf{96.3}\% & \diffcell{+36.7}
             & 41.9\% & \textbf{93.7}\% & \diffcell{+51.8} \\
           \bottomrule
       \end{tabular}
\end{table*}

\paragraph{Metrics.}
We evaluate the realized schedule using the following metrics:
\begin{itemize}
    \item \textbf{Quality-pass rate.} A schedule passes if an independent checker confirms validity, including any submitted hard relations, and its makespan is at most 1.25 times the known optimum. For task $\tau$, let $C_{\max}(y_\tau)$ denote the makespan of schedule $y_\tau$ and $C^\star_\tau$ the known optimum. We report
    \begin{multline}
        QPR=\frac{1}{|\mathcal{T}|}
        \sum_{\tau\in\mathcal{T}}\mathbf{1}[
            V(y_\tau)=1,\\
            \text{ and }
            C_{\max}(y_\tau)\leq1.25C^\star_\tau].
    \end{multline}
    Known optima and quality-pass labels are used for evaluation only, and hence are not exposed to the agent.
    \item \textbf{First-proposal quality-pass rate.} We apply the same pass rate criterion to the first proposal to measure whether the interface delivers a usable schedule before revision.
\end{itemize}

\paragraph{Baselines and controls.}
We compare \system with \textbf{Code (C1T)}, a baseline in which the agent returns Python source code. 
The sandboxed program outputs either relations for the shared backend or directly computed operation start times. 
This comparison tests code-mediated versus direct relation delivery while preserving access to the same backend. 
A separate model-free control submits the input job order with no additional hard relations, evaluating backend capability independently of model-generated actions.

\paragraph{Instantiating \system for scheduling tasks.}
We instantiate the contract with $\mathcal{O}$ exposing jobs, operation durations, machine assignments, and the relation schema.
The action space $\mathcal{A}$ accepts job priorities and named operation relations, assigning timestamp construction to the bounded scheduling backend $E$.
Invalid or cyclic relations are rejected rather than silently repaired.
An independent verifier $V$ checks the realized schedule and preservation of hard relations.
Feedback $F$ returns validity, makespan, and violation evidence; valid schedules additionally expose their start times for the next decision.
The agent can therefore revise its scheduling preferences while the backend remains responsible for their temporal realization.

\subsection{Experimental Results}

\paragraph{\system improves scheduling action delivery.}
Table~\ref{tab:scheduling-main-quality} compares quality-pass rates across all models.
\system reaches 260/270 quality passes (96.3\%), compared with 161/270 (59.6\%) for code-backed delivery.
These results show that \system improves the pass rate for seven models and ties for two, reaching 100.0\% for seven models.

\paragraph{\system delivers usable schedules earlier.}
The first proposal passes on 253/270 episodes (93.7\%) with \system, compared with 113/270 (41.9\%) with code-backed delivery.
Thus, \system's first-proposal pass rate already exceeds the baseline's final rate after three proposals.

\paragraph{Additional results.}
We show that relation improves access to backend capability in Appendix \ref{app:scheduling-results}.

\section{Conclusion}
\label{sec:conclusion}

We introduced \system, a relational agent contract for constraint-governed engineering tasks. 
\system makes the agent--tool boundary explicit: agents observe authoritative state and retain semantic engineering judgment, while specialized tools derive mechanically determined consequences, realize the artifact, and expose it to independent verification. 
This separation allows an agent to decide what an artifact should satisfy without requiring it to reproduce the precise geometry or target-specific implications that engineering tools can derive more reliably.
Our evaluation on nine models shows that \system achieves the highest clean-pass rate for most of the models. 
Moreover, \system reduces token use for most models and lowers latency.
Our results establish interface design as a first-class source of engineering-agent capability. 

\section*{Limitations}
\label{sec:limitations}

\system moves derivation from the agent to domain tools, so it applies only where a realizer $E$ and an independent checker $V$ exist or can be built.
Choosing or building these components is a nontrivial engineering cost, and the results of \system are bounded by their coverage and correctness.
We count an artifact as successful when it satisfies every predicate in $\Phi$, as judged by our checkers.
Requirements that cannot be written as checkable predicates, such as loosely specified preferences, fall outside our formulation.

\section*{Ethical Considerations}
This study evaluates AI agents in constraint-governed engineering tasks. 
It involves no human participants or collection of personal data. 
Automatically generated engineering artifacts may have physical consequences if used outside a research setting. 
Such use requires appropriate engineering review and hardware validation.

\bibliography{references}

\appendix
\newpage
\section*{Appendix}

\section{Additional PCB Analyses}
\label{app:additional-pcb-results}

The PCB analyses first test whether committed relations survive geometric changes. Later sections examine action delivery in scheduling and embedded-code generation, show recorded PCB examples, and test protocol reuse with deterministic cases.

\subsection{Robustness to board and footprint changes}
\label{sec:portability}

When board or component dimensions change, the same committed relations must be realized in new geometry. We test whether \system can reuse its model-authored relations in this setting by holding those relations fixed and comparing adaptation methods.  This isolates the value of retaining relational decisions after the original model call.

\subsubsection{Experimental Setup}

\paragraph{Agents and tasks.}
We use the same agents as in Section \ref{sec:setup}.
Each agent generates an initial artifact for 26 synthetic placement families, giving 234 initial attempts.  
Of these, 153 pass the original full checker.
We then apply three deterministic changes to each passing artifact: a 20\% board shrink, a 20\% board expansion, and a 25\% footprint shrink.  
This produces $153\times3=459$ adapted layouts.

\paragraph{Metrics and statistical analysis.}
We assess adapted layouts with the clean-pass criterion in Section~\ref{sec:setup} and report two rates.  
The adaptation rate considers the 459 changes to initially successful layouts.  
The overall rate considers all $234\times3=702$ attempted changes, counting an initial generation failure as a failure under each change. 
Both calculations include the 24 board-shrink cases whose feasibility remains unestablished.  
For differences between adaptation methods, we resample the 26 placement families 4,000 times, keeping results from each family together, to form descriptive 95\% confidence intervals without multiple-comparison adjustment.  
Results use this study's preselected seed 0; seeds 1 and 2 test sensitivity.

\paragraph{Baselines and controls.}
All adaptation methods start from the same successful \system-produced artifact, so their comparison concerns how the initial layout responds to geometric change.
\begin{itemize}
\item \textbf{Frozen coordinates:} retain the original component locations.
\item \textbf{Affine coordinates:} scale component locations with the board.
\item \textbf{Empty-relation solver:} invoke the same placement solver without the original relations.
\item \textbf{Preserve-legal coordinates:} use affine coordinates when mechanically legal and otherwise invoke the empty-relation solver.
\item \textbf{Numeric best-of-three:} select a passing outcome, if available, among frozen coordinates, affine coordinates, and the empty-relation solver using the final checker.  This control shows the best result obtainable from those three methods with hindsight.
\item \textbf{\system relational re-solve:} invoke the same placement solver with the unchanged \system-generated relations to recompute coordinates in the transformed context.
\end{itemize}

\subsubsection{Experimental Results}

\begin{table*}[t]
    \centering
    \caption{Average token use per episode on PCB tasks. \system improves token efficiency.}
    \label{tab:construction-tokens}
    \setlength{\tabcolsep}{4.5pt}
    \renewcommand{\arraystretch}{1.12}
    \newcommand{\tokcell}[1]{#1k}
    \begin{tabular}{@{}lcccccc@{}}
        \toprule
        Model & Coord. & Coord.+Py & Candidates & \shortstack{Seq.\\Relations} & Req.-Only & \textbf{\system} \\
        \midrule
        \leaderlogo{assets/logos/Deepseek.png}\ DeepSeek-V4-Flash
          & \tokcell{7.5} & \tokcell{8.8} & \tokcell{10.2} & \tokcell{8.8} & \tokcell{\textbf{6.6}} & \tokcell{\textbf{6.6}} \\
        \leaderlogo{assets/logos/glm.png}\ GLM-5.3-Flash
          & \tokcell{11.0} & \tokcell{11.8} & \tokcell{12.8} & \tokcell{10.5} & \tokcell{9.3} & \tokcell{\textbf{8.8}} \\
        \leaderlogo{assets/logos/openai_logo.png}\ GPT-5.6-Luna
          & \tokcell{9.9} & \tokcell{10.1} & \tokcell{11.4} & \tokcell{9.3} & \tokcell{7.2} & \tokcell{\textbf{6.0}} \\
        \leaderlogo{assets/logos/openai_logo.png}\ GPT-5.6-Sol
          & \tokcell{9.4} & \tokcell{9.7} & \tokcell{11.1} & \tokcell{9.2} & \tokcell{7.6} & \tokcell{\textbf{6.1}} \\
        \leaderlogo{assets/logos/moonshot.png}\ Kimi-K2.7
          & \tokcell{10.0} & \tokcell{9.6} & \tokcell{10.5} & \tokcell{11.0} & \tokcell{8.7} & \tokcell{\textbf{7.2}} \\
        \leaderlogo{assets/logos/qwen.png}\ Qwen3.6-27B
          & \tokcell{36.6} & \tokcell{43.2} & \tokcell{25.8} & \tokcell{20.6} & \tokcell{14.0} & \tokcell{\textbf{13.3}} \\
        \leaderlogo{assets/logos/claude_flower.png}\ Claude-Opus-4.8
          & \tokcell{10.2} & \tokcell{11.1} & \tokcell{12.9} & \tokcell{11.6} & \tokcell{\textbf{9.7}} & \tokcell{9.9} \\
        \leaderlogo{assets/logos/claude_flower.png}\ Claude-Opus-5
          & \tokcell{12.4} & \tokcell{13.3} & \tokcell{14.0} & \tokcell{12.4} & \tokcell{11.1} & \tokcell{\textbf{10.6}} \\
        \leaderlogo{assets/logos/gemini.png}\ Gemini-3.8-Flash
          & \tokcell{31.1} & \tokcell{21.9} & \tokcell{18.0} & \tokcell{18.3} & \tokcell{14.9} & \tokcell{\textbf{13.0}} \\
        \bottomrule
    \end{tabular}
\end{table*}

\paragraph{\system preserves relations under geometric perturbations.}
Relational re-solving passes 411/459 adapted layouts (89.5\%), compared with 301/459 (65.6\%) for the numeric best-of-three (Table~\ref{tab:portability}).  The paired difference is $+24.0$ percentage points, with a 95\% interval of $[14.1,34.5]$ after grouping results by placement family.  Including failed initial generations gives overall rates of 411/702 (58.5\%) and 301/702 (42.9\%), respectively.  Retaining the original relations thus improves adaptation over controls that retain, scale, or regenerate coordinates without them.

\begin{table*}[!ht]
\centering
\caption{Robustness to board and footprint changes. Adaptation rates cover 459 changes to initially successful layouts; overall rates include all 702 attempted changes. Differences compare relational re-solving with each control, in percentage points, with descriptive 95\% bootstrap intervals that group results by placement family.}
\label{tab:portability}
\small
\setlength{\tabcolsep}{5.0pt}
\begin{tabular}{@{}lccc@{}}
\toprule
Adaptation method & Adaptation clean-pass & Overall clean-pass & Paired difference [95\% CI] \\
\midrule
Frozen coordinates & 187/459 (40.7\%) & 187/702 (26.6\%) & \diffcell{$+48.8$} [$37.8$, $60.7$] \\
Affine coordinates & 239/459 (52.1\%) & 239/702 (34.0\%) & \diffcell{$+37.5$} [$28.0$, $46.6$] \\
Empty-relation solver & 212/459 (46.2\%) & 212/702 (30.2\%) & \diffcell{$+43.4$} [$27.1$, $59.8$] \\
Preserve-legal coordinates & 267/459 (58.2\%) & 267/702 (38.0\%) & \diffcell{$+31.4$} [$20.6$, $42.2$] \\
Numeric best-of-three & 301/459 (65.6\%) & 301/702 (42.9\%) & \diffcell{$+24.0$} [$14.1$, $34.5$] \\
\system relational re-solve & \textbf{411/459 (89.5\%)} & \textbf{411/702 (58.5\%)} & --- \\
\bottomrule
\end{tabular}
\end{table*}

\paragraph{The robustness gain holds across all three perturbations.}
The advantage holds for each perturbation (Table~\ref{tab:portability-transforms}).  Relational re-solving passes 77.1\% of board-shrink cases, 92.2\% of board-expansion cases, and 99.3\% of footprint-shrink cases.  Its largest gap over the numeric best-of-three is under board shrink (77.1\% versus 47.7\%).  In each case, the relations stay fixed while the backend recomputes coordinates for the changed geometry.

\begin{table*}[!ht]
\centering
\caption{Clean-pass rates by geometric change, among the 153 initially successful layouts.}
\label{tab:portability-transforms}
\small
\begin{tabular}{@{}lrrr@{}}
\toprule
Transform & Frozen & Numeric best-of-three & Relational re-solve \\
\midrule
Board shrink 20\% & 23/153 (15.0\%) & 73/153 (47.7\%) & \textbf{118/153 (77.1\%)} \\
Board expansion 20\% & 46/153 (30.1\%) & 110/153 (71.9\%) & \textbf{141/153 (92.2\%)} \\
Footprint shrink 25\% & 118/153 (77.1\%) & 118/153 (77.1\%) & \textbf{152/153 (99.3\%)} \\
\bottomrule
\end{tabular}
\end{table*}

\paragraph{Relational reuse is stable across solver seeds.}
The result changes little across solver seeds: relational re-solving passes 411/459 adapted layouts at seed 0, 410/459 at seed 1, and 408/459 at seed 2.  Across models, adaptation success ranges from 18/24 (75.0\%) to 58/60 (96.7\%); each model starts from 8--23 successful initial artifacts, which undergo 24--69 changes.  Across the tested changes and seeds, the same relations remain reusable without another model decision.

\subsection{Example Tasks and Model Outputs}
\label{sec:pcb-examples}

We present three sample tasks in \ref{tab:pcb-task-examples} from placement, routing, and combined categories.
The GPT-5.6-Luna outputs on these sample tasks were generated with \system and seed 1, as described in Section~\ref{sec:setup}. 

\begin{table*}[!ht]
\centering
\caption{Recorded PCB examples. All three realized artifacts pass their task-specific relation checks, complete routing, and zero-design-rule-violation checks.}
\label{tab:pcb-task-examples}
\small
\begin{tabularx}{\textwidth}{@{}lXX@{}}
\toprule
Task & Requirement & Verification of realized artifact \\
\midrule
P013: placement & Associate connector J1 with the left board edge and J2 with the right edge. & Each connector is 1.00\,mm from its requested edge, meeting the task's 1.00\,mm limit. \\
R010: routing & Match the routed lengths of nets N2 and N3. & The measured mismatch is 0.00\,mm, within the task oracle's 1.00\,mm tolerance. \\
C008: combined & Place J1 at the top edge, decoupling capacitors C1--C3 near U1 pins, and preserve the R1/C4 and R2/C5 filter groups. & J1 is 1.00\,mm from the top edge; capacitor-to-pin distances are 2.00, 1.50, and 1.58\,mm; filter-group spreads are 2.69 and 3.04\,mm. \\
\bottomrule
\end{tabularx}
\end{table*}

\paragraph{Placement through named relations.}
  For P013, the model names the connectors and their requested board sides:
  \begin{quote}\small
  \begin{verbatim}
  placement:
    board_edge(J1, left)
    board_edge(J2, right)
  routing: []
  \end{verbatim}
  \end{quote}
  The submission specifies no coordinates or routing directives. The backend places both connectors and routes the required
  connections. On the realized board, the independent checker measures each connector 1.00\,mm from its requested edge, meeting the
  task's 1.00\,mm limit, and confirms complete routing with zero design-rule violations.

  \paragraph{Routing through a categorical matching requirement.}
  For R010, the model names the nets and a tolerance class:
  \begin{quote}\small
  \begin{verbatim}
  placement: []
  routing:
    route_matched([N2, N3], 5%)
  \end{verbatim}
  \end{quote}
  The backend determines the copper paths and reports a 15.0\,mm target length. Independent measurement finds a 0.00\,mm mismatch,
  within the task's 1.00\,mm tolerance. Numeric path lengths appear in the execution evidence, not in the model's submission.

  \paragraph{Joint relations for coupled placement and routing.}
  For C008, the model submits eight placement relations and no explicit routing directive:
  \begin{quote}\small
  \begin{verbatim}
  placement:
    board_edge(J1, top)
    place_relative(C1, U1.R1, near, tight)
    place_relative(C2, U1.R2, near, tight)
    place_relative(C3, U1.R3, near, tight)
    place_relative(R1, U1.R4, near, close)
    place_relative(C4, R1, near, close)
    place_relative(R2, U1.R5, near, close)
    place_relative(C5, R2, near, close)
  routing: []
  \end{verbatim}
  \end{quote}
  The backend realizes these relations while routing all required nets. Independent checks place J1 1.00\,mm from the top edge, measure
  C1--C3 at 2.00, 1.50, and 1.58\,mm from their respective pins, and measure spreads of 2.69 and 3.04\,mm for the two filter groups.
  The same realized board satisfies the placement requirements, complete connectivity, and zero-design-rule-violation check.

\section{Scheduling Action Delivery}
\label{app:scheduling-results}

\subsection{Experimental Setup}

\paragraph{Agents and tasks.}
Nine models each produce three successive proposals for every scheduling instance from LA01 to LA30 under both interfaces.  This yields 270 direct comparisons on the same model and instance.

\paragraph{Metrics and statistical analysis.}
Quality pass requires independent schedule validity and a makespan no greater than 1.25 times the known optimum.  We compare the final and first proposals from both interfaces and estimate a 95\% interval for the difference in final pass rates by resampling the 30 instances 5,000 times.

\paragraph{Baselines and controls.}
\system returns relation directly: a soft permutation of all jobs plus up to eight hard end-before-start operation pairs.  \textbf{C1T} returns Python source in a JSON \code{code} field; the program may print the same relations or direct start times.  When either interface returns relations, it invokes the same bounded backend with 8,192 candidate evaluations, eight restarts, and seed zero.  The comparison therefore tests action delivery through the two interfaces with a shared relation backend.

\subsection{Experimental Results}

\paragraph{Direct relational actions improve quality pass.}
\system achieves final quality pass in 260/270 runs (96.3\%), compared with 161/270 (59.6\%) for C1T, a difference of $+36.7$ percentage points (95\% interval $[31.1,41.9]$). 
On the same model and instance, both interfaces pass 156 times; \system alone passes 104 times, C1T alone five times, and neither passes 34 times. 
\system also produces a passing first proposal in 253/270 runs (93.7\%), compared with 113/270 (41.9\%) for C1T.  
Each run still uses all three model calls, so the first-proposal result shows earlier usable action.

\paragraph{Direct relations make backend capability accessible to agents.}
  A fixed, model-free relation set produces a quality-passing schedule in all 40 calibration checks on LA11--LA30 across the two interfaces. 
  This confirms that both interfaces can access a capable shared backend. In the model-driven comparison, however, direct relational submission achieves 260/270 quality passes, compared with 161/270 through code-mediated delivery. 
  The gain therefore reflects more reliable delivery of agent's relations to the backend, rather than a difference in backend capability.

 \subsection{Example Task and Model Output}
LA07 contains 15 jobs and five machines. In its final proposal, GPT-5.6-Sol retains its job priorities and requires operation \texttt{j2o4} to precede \texttt{j4o4}:
  \begin{quote}\footnotesize
  \begin{verbatim}
priority_jobs: [8, 1, 14, 13, 0, 10, 7, 5,
                3, 6, 2, 12, 9, 4, 11]
before: [(j2o4, j4o4)]
\end{verbatim}
  \end{quote}
The independent checker confirms the hard relation and schedule validity.
The realized makespan is 890, matching the known optimum; the first proposal, which omitted the relation, had makespan 892.

\section{Evaluation of \system on Embedded-Code Generation Tasks}
\label{app:embedded-results}

\subsection{Experimental Setup}

\paragraph{Agents and tasks.}
We evaluate \system on embedded-code generation as a constraint-governed engineering task: a program must implement the requested behavior while satisfying obligations imposed by its microcontroller and board. 
These obligations can couple implementation choices to memory reachability, cache maintenance, interrupt use, peripheral clocking, and pin assignments; the same requested behavior may therefore require different programs on different targets. 
The evaluation comprises 141 tasks---121 adapted from EmbedEval~\citep{embedeval2026} and 20 newly authored hardware-obligation tasks---assigned to four microcontroller targets: nRF52840, STM32H743, ESP32-S3, and STM32F407. 
We evaluate eight models: DeepSeek-V4-Flash, Qwen3.6-27B, GLM-5.3-Flash, Kimi-K2.7, GPT-5.6-Luna, GPT-5.6-Sol, Claude-Opus-5, and Claude-Opus-4.8.

\paragraph{Metrics and statistical analysis.}
Each run has a recorded outcome of pass, fail, or unknown. 
For an interface with $P$ passes and $U$ unknown outcomes among $N$ runs, we report the pass-rate range $[P/N,\,(P+U)/N]$: its endpoints count every unknown outcome as a failure or as a pass, respectively. 
These endpoints are bounds on the recorded pass rate. 

\paragraph{Baselines and controls.}
We consider three source-writing controls as alternative interfaces. 
For each task, all interfaces use the same requested behavior, target, and public platform state. 
The three source-writing controls vary how task-relevant public SDK information is presented. \textbf{RawSDK} provides excerpts from the SDK sources, including declarations and their documentation. 
\textbf{SignaturesOnly} provides the selected declarations without explanatory documentation. 
\textbf{SourceLite} presents the selected declarations and non-boilerplate documentation clauses in a compact, structured form. 
Under each of these controls, the model must produce the requested source or configuration artifact.
\system instead asks the model for a typed semantic plan specifying the operation and its parameters; a deterministic tool validates and lowers that plan into the implementation artifact. 
Thus, the controls vary the context available for model-authored implementation, while \system changes which parts of implementation the model must produce.

\subsection{Experimental Results}

\paragraph{\system improves pass rates and token use.}
Pass-rates and token usage counts are presented in Table \ref{tab:embedded-pass-by-model} and \ref{tab:embedded-tokens-by-model}.
Across all runs, \system has 827 passes, 293 failures, and eight unknown outcomes. 
Its pass-rate bound is 73.3--74.0\%, compared with 46.7--53.1\% for SourceLite.  
Among the 998 directly comparable cases, \system's pass rate is 21.2 percentage points higher. 
\system also uses 39.0\% fewer tokens in these cases. 
Across all four interfaces, 224 of the 4,512 outcomes are unknown.
\begin{table*}[t]
      \centering
      \caption{Pass rates by model on complete four-interface pairs. $n$ is the number of tasks for that model with known outcomes and
      token counts under every interface; the pooled row contains 998 model--task pairs. The best pass rate in each row is bold.}
      \label{tab:embedded-pass-by-model}
      \small
      \begin{tabular}{@{}lrrrrr@{}}
          \toprule
          Model & $n$ & RawSDK & SignaturesOnly & SourceLite & \system \\
          \midrule
          \leaderlogo{assets/logos/Deepseek.png}\ DeepSeek-V4-Flash
              & 132 & 43.9\% & 39.4\% & 42.4\% & \textbf{68.2\%} \\
          \leaderlogo{assets/logos/qwen.png}\ Qwen3.6-27B
              & 129 & 39.5\% & 33.3\% & 41.9\% & \textbf{60.5\%} \\
          \leaderlogo{assets/logos/glm.png}\ GLM-5.3-Flash
              & 106 & 41.5\% & 38.7\% & 40.6\% & \textbf{70.8\%} \\
          \leaderlogo{assets/logos/moonshot.png}\ Kimi-K2.7
              & 130 & 43.8\% & 46.2\% & 44.6\% & \textbf{63.1\%} \\
          \leaderlogo{assets/logos/openai_logo.png}\ GPT-5.6-Luna
              & 140 & 59.3\% & 55.7\% & 57.9\% & \textbf{87.1\%} \\
          \leaderlogo{assets/logos/openai_logo.png}\ GPT-5.6-Sol
              & 121 & 62.8\% & 59.5\% & 66.1\% & \textbf{78.5\%} \\
          \leaderlogo{assets/logos/claude_flower.png}\ Claude-Opus-5
              & 115 & 39.1\% & 44.3\% & 45.2\% & \textbf{66.1\%} \\
          \leaderlogo{assets/logos/claude_flower.png}\ Claude-Opus-4.8
              & 125 & 65.6\% & 60.8\% & 67.2\% & \textbf{81.6\%} \\
          \midrule
          Pooled & 998 & 49.7\% & 47.4\% & 50.9\% & \textbf{72.1\%} \\
          \bottomrule
      \end{tabular}
  \end{table*}

  \begin{table*}[t]
      \centering
      \caption{Model input and output tokens on the same complete pairs as Table~\ref{tab:embedded-pass-by-model}. Each entry sums
      tokens over the model's $n$ paired tasks; tool-generated code is excluded. The lowest token total in each row is bold.}
      \label{tab:embedded-tokens-by-model}
      \small
      \begin{tabular}{@{}lrrrrr@{}}
          \toprule
          Model & $n$ & RawSDK & SignaturesOnly & SourceLite & \system \\
          \midrule
          \leaderlogo{assets/logos/Deepseek.png}\ DeepSeek-V4-Flash
              & 132 & 894,940 & 640,572 & 846,254 & \textbf{525,295} \\
          \leaderlogo{assets/logos/qwen.png}\ Qwen3.6-27B
              & 129 & 1,478,250 & 1,198,840 & 1,405,793 & \textbf{858,130} \\
          \leaderlogo{assets/logos/glm.png}\ GLM-5.3-Flash
              & 106 & 762,785 & 575,674 & 725,054 & \textbf{446,729} \\
          \leaderlogo{assets/logos/moonshot.png}\ Kimi-K2.7
              & 130 & 765,991 & 538,520 & 726,882 & \textbf{468,364} \\
          \leaderlogo{assets/logos/openai_logo.png}\ GPT-5.6-Luna
              & 140 & 944,558 & 701,345 & 906,761 & \textbf{514,919} \\
          \leaderlogo{assets/logos/openai_logo.png}\ GPT-5.6-Sol
              & 121 & 817,766 & 616,651 & 784,798 & \textbf{476,405} \\
          \leaderlogo{assets/logos/claude_flower.png}\ Claude-Opus-5
              & 115 & 1,337,597 & 1,010,766 & 1,270,340 & \textbf{756,468} \\
          \leaderlogo{assets/logos/claude_flower.png}\ Claude-Opus-4.8
              & 125 & 1,267,747 & 894,642 & 1,188,824 & \textbf{748,507} \\
          \midrule
          Pooled & 998 & 8,269,634 & 6,177,010 & 7,854,706
                 & \textbf{4,794,817} \\
          \bottomrule
      \end{tabular}
  \end{table*}

\subsection{Examples}
 \paragraph{Periodic ADC sampling.}
  Task adc-001 asks for a Zephyr application that reads a devicetree ADC
  channel every second and prints the result in millivolts. The model submits:
  \begin{quote}\small
  \begin{verbatim}
  target: nrf52840
  operation: adc_dt(
    path=zephyr_user, channel_index=0,
    interval_ms=1000, convert_mv=true)
  \end{verbatim}
  \end{quote}
  The plan identifies the device binding, channel, sampling interval, and
  conversion requirement. The backend lowers it to C source; the recorded
  artifact passes the task's native-compilation and static checks.

  \paragraph{PWM LED fading.}
  Task pwm-001 requires an LED duty cycle that rises and falls in
  2{,}000{,}000\,ns steps within a 20{,}000{,}000\,ns period, pausing
  50\,ms between steps. The model submits:
  \begin{quote}\small
  \begin{verbatim}
  target: nrf52840
  operation: pwm_triangle(
    led=pwm_led0, period_ns=20000000,
    step_ns=2000000, interval_ms=50,
    ramp_down=true)
  \end{verbatim}
  \end{quote}
  The submission specifies the fade behavior without writing the Zephyr
  control loop. The lowered artifact passes the recorded native-compilation
  and static checks.

\section{Cross-Domain \system Instantiation Study}
  \label{sec:protocol-reuse}

  We test whether responsibility separation works in an executable implementation by adding a bounded Boolean-logic task to the unchanged \system runtime. 
  The question is whether a small binding can connect the new task's observations, actions, realization, verification, and feedback to the shared turn sequence while preserving the behavior of existing domains.

  \subsection{Experimental Setup}

  \paragraph{The logic task.}
  Each task specifies binary input and output ports and a complete truth table: for every possible assignment of input bits, the table
  states the required output bits. The requested artifact is a \emph{combinational Boolean circuit}. 
  Here, a circuit means an abstract, acyclic network of Boolean operations (AND, OR, NOT, and constants). 
  Its outputs depend only on the current inputs; it has no memory or clock. 
  In this study, \emph{logic synthesis} means constructing that network from a
  supplied action. The action lists, for each output, the truth-table rows on which that output should equal one; these row numbers are
  called minterm indices. We use ten constructed designs with at most four inputs, so every input assignment can be checked.

  \paragraph{Domain implementation.}
  A deterministic realizer uses SymPy's sum-of-products procedure to turn the supplied minterm indices into a Boolean network
  represented as a directed acyclic graph. A separate pure-Python verifier checks the graph's structure and evaluates it on every input assignment against both the required truth table and the supplied action. The verifier imports neither SymPy nor the synthesis module and does not trust the realizer's receipt.

  The \system runtime orders five operations for each proposal: present an observation, validate the action, realize an artifact, verify it, and return feedback. A lightweight domain binding connects the logic implementation to these operations; the shared runtime is unchanged. 

  \paragraph{Checks and comparisons.}
  For each design, we run four predefined checks: a correct action, an action changed on one truth-table row, a modified artifact
  accompanied by a forged success receipt, and a supplied two-step repair trace that first fails and then passes. These 40 checks use
  supplied actions and repairs; they do not involve model calls. We also compare \system with a direct driver that invokes the same five domain operations in the same order. Both paths receive the same supplied actions and use the same realizer and verifier. The
  comparison covers 400 paired logic executions and 72 paired executions from PCB, scheduling, and embedded-code tasks. 
  A separate compatibility check replays 128 outcomes from those three domains.

  \subsection{Experimental Results}

  \paragraph{Logic calibration.}
  Results from all 40 checks show that the verifier accepts every correct action, rejects every one-row change, and rejects every modified artifact despite its forged receipt. 
  Each supplied repair trace changes from failure to success. Across the ten designs, exhaustive verification covers 92 input assignments and 208 output bits per complete positive cohort.

  \begin{table*}[!ht]
      \centering
      \caption{Bounded logic calibration using supplied actions and repairs, with no model calls.}
      \label{tab:logic-calibration}
      \small
      \begin{tabular}{@{}lr@{}}
          \toprule
          Item & Verified value \\
          \midrule
          Constructed designs & 10 \\
          Assignments / output bits exhaustively checked & 92 / 208 \\
          Correct actions accepted & 10/10 \\
          One-row action changes rejected & 10/10 \\
          Modified artifacts with forged receipts rejected & 10/10 \\
          Supplied repair traces, fail $\rightarrow$ pass & 10/10 \\
          Domain binding footprint & 30 physical / 22 code lines \\
          Full logic-domain implementation & 191 physical / 155 code lines \\
          \bottomrule
      \end{tabular}
  \end{table*}

 \begin{table*}[!ht]
      \centering
      \caption{Paired execution through the direct driver and \system. Outputs, callbacks, and message bytes match for every trace.
      Paired $\Delta$ is the median over traces of the per-trace runtime difference (\system minus direct), so it can differ in value
      and sign from the difference between the two medians. Overhead is the percentage difference in total runtime,
      $(T_{\system}-T_{\text{direct}})/T_{\text{direct}}$, with a descriptive 95\% bootstrap interval that resamples by test case.}
      \label{tab:controller-runtime}
      \footnotesize
      \setlength{\tabcolsep}{3.2pt}
\begin{tabular}{@{}lrrrrrrr@{}}
          \toprule
          Domain & Traces & Matching outputs & Direct median (ms) & \system median (ms) & Paired $\Delta$ (ms) & Overhead & 95\% interval \\
          \midrule
          Logic      & 400 & 400/400 & 1.211    & 1.658    & \diffcell{$+$0.398}     & \diffcell{$+$30.84\%}   & [$27.57$, $34.94$] \\
          PCB        & 24  & 24/24   & 59.737   & 59.451   & \diffcellneg{$-$0.158}  & \diffcellneg{$-$0.66\%} & [$-1.90$, $0.36$] \\
          Scheduling & 24  & 24/24   & 2370.016 & 2372.419 & \diffcellneg{$-$0.304}  & \diffcellneg{$-$0.24\%} & [$-1.01$, $0.56$] \\
          Embedded   & 24  & 24/24   & 4011.018 & 3940.803 & \diffcell{$+$38.725}    & \diffcell{$+$4.21\%}    & [$-1.46$, $10.18$] \\
          \bottomrule
      \end{tabular}
  \end{table*}

  \paragraph{Behavior and runtime.}
  For all 400 logic executions and 72 executions from the existing domains, \system and the direct driver produce identical artifacts,
  verdicts, callback counts, and serialized messages. Table~\ref{tab:controller-runtime} reports runtime on the same host after warm-
  up. On the logic task, \system adds a median of 0.398\,ms to a native path of roughly 1.2\,ms. The runtime intervals for PCB, scheduling, and embedded code all include zero.

  \paragraph{Compatibility with existing domains.}
  The compatibility check reproduces all 52 PCB, 60 scheduling, and 16 embedded-code outcomes, including 15 recorded failures.
  Thus, the unchanged turn implementation accommodates this additional, independently checked artifact type while preserving the
  supplied interactions in the existing domains. 
  The logic implementation also makes the division of work concrete: the binding to \system is small, while the action format, realizer, and verifier remain specific to the engineering task.

\section*{Acknowledgement}
In this work, we used generative AI tools for implementing and running experiments, finding related work, cleaning and reformating dataset, and improving the clarity of the manuscript. 
We have not used generative AI tools for research ideation, assisting with translation, supporting qualitative and thematic data analysis, or interpreting results.
We have reviewed all AI-assisted work including code, paper edit suggestions, search results. 
We take responsibility for the final content of this work, including text, claims or artifacts produced with the aid of generative AI.

\end{document}